\documentclass[letterpaper, 10 pt, conference]{ieeeconf}  % Comment this line out if you need a4paper

\IEEEoverridecommandlockouts                              % This command is only needed if 
\usepackage{graphics} % for pdf, bitmapped graphics files
\usepackage{amsmath} % assumes amsmath package installed
\usepackage{amsmath,graphicx}
\usepackage{xcolor}

\title{\LARGE \bf
Face Density as a Proxy for Data Complexity: \\ Quantifying the Hardness of Instance Count
}

\author{Abolfazl Mohammadi-Seif and Ricardo Baeza-Yates%
\thanks{This work has been partially supported by MCIN/AEI/10.13039/501100011033 under the Mar\'ia de Maeztu Units of Excellence Program (CEX2021-001195-M).}%
\thanks{The authors are with the Department of Engineering, Universitat Pompeu Fabra, Barcelona, Spain.
        {\tt\small abolfazl.mohammadiseif@upf.edu, rbaeza@acm.org}.}%
\thanks{This work has been accepted for publication in the Proceedings of IEEE CAI 2026. The final published version should be cited.}%
}

\begin{document}

\maketitle
\thispagestyle{empty}
\pagestyle{empty}

%%%%%%%%%%%%%%%%%%%%%%%%%%%%%%%%%%%%%%%%%%%%%%%%%%%%%%%%%%%%%%%%%%%%%%%%%%%%%%%%
\begin{abstract}
Machine learning progress has historically prioritized model-centric innovations, yet achievable performance is frequently capped by the intrinsic complexity of the data itself. In this work, we isolate and quantify the impact of instance density (measured by face count) as a primary driver of data complexity. Rather than simply observing that ``crowded scenes are harder,'' we rigorously control for class imbalance to measure the precise degradation caused by density alone.

Controlled experiments on the \textit{WIDER FACE} and \textit{Open Images} datasets, restricted to exactly 1 to 18 faces per image with perfectly balanced sampling, reveal that model performance degrades monotonically with increasing face count. This trend holds across classification, regression, and detection paradigms, even when models are fully exposed to the entire density range. Furthermore, we demonstrate that models trained on low-density regimes fail to generalize to higher densities, exhibiting a systematic under-counting bias, with error rates increasing by up to 4.6$\times$, which suggests density acts as a domain shift. 

These findings establish instance density as an intrinsic, quantifiable dimension of data hardness and motivate specific interventions in curriculum learning and density-stratified evaluation.
\end{abstract}
%%%%%%%%%%%%%%%%%%%%%%%%%%%%%%%%%%%%%%%%%%%%%%%%%%%%%%%%%%%%%%%%%%%%%%%%%%%%%%%%

\section{Introduction}
\label{sec:intro}

The past decade of machine learning has been dominated by model-centric innovation: ever-larger architectures, sophisticated optimization techniques, massive pre-training, and clever regularization schemes have driven remarkable empirical gains across standard benchmarks. Yet, despite this relentless focus on improving models, real-world performance plateaus persist in many domains, particularly in computer vision tasks involving crowded scenes, multi-object interaction, or heavy occlusion. These failures are frequently attributed to insufficient model capacity, suboptimal hyperparameters, or lack of more data. However, a growing body of evidence suggests a more fundamental culprit: the intrinsic complexity of the instance data itself.

Although model improvements can push performance upward, they cannot exceed the ceiling imposed by the hardness of the instance complexity. When instances heavily overlap, vary dramatically in scale, or crowd the visual field, even the most powerful architectures struggle, not because they lack capacity, but because the problem itself becomes harder. This phenomenon is not unique to vision; it appears wherever instance density or interaction complexity grows (e.g., long-sequence modeling in NLP, multi-agent prediction in robotics, or dense point clouds in 3D perception), yet it remains underexplored compared to architectural advances.

\begin{figure}[t]
  \centering
  \includegraphics[width=\columnwidth]{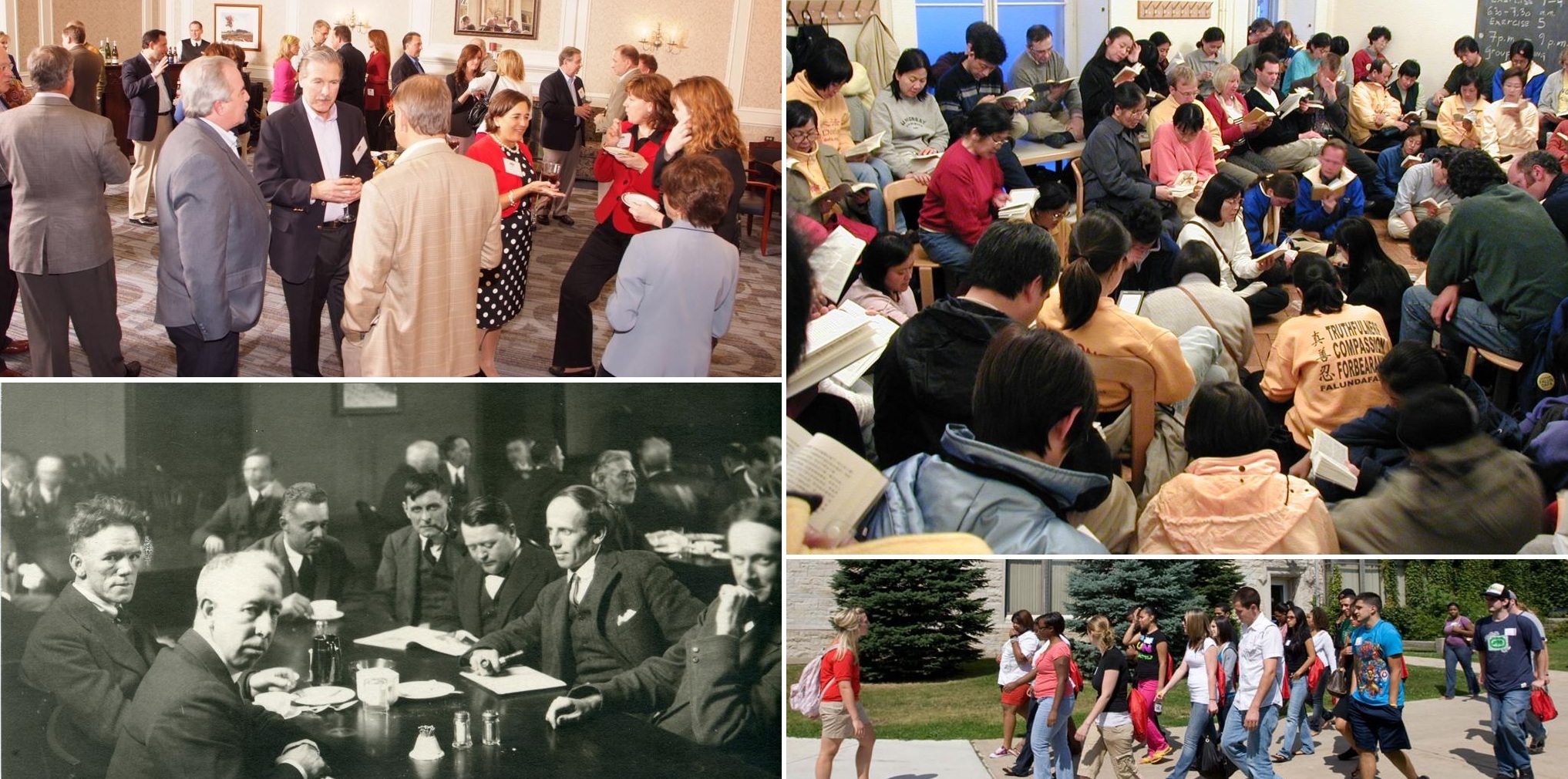}
  \caption{Visualizing Instance Complexity (WIDER FACE): Increasing density creates intrinsic difficulties (occlusion, scale, entanglement) regardless of model capacity.}
  \label{fig:intro_complexity}
\end{figure}

In this work, we deliberately shift the perspective from model-centric to data-centric by formalizing \textit{Instance Complexity} through the proxy of \textit{Instance Density}, specifically, the number of faces per image. We argue that density is not merely a contextual feature but a quantifiable dimension of hardness that imposes a performance ceiling independent of model capacity. The face count is clean, objective, perfectly controllable, and directly tied to real emergent difficulties (occlusion, scale variation, spatial crowding, and feature entanglement, see Fig.~\ref{fig:intro_complexity}) without requiring complex annotations of those individual factors.

To ensure the robustness and generalizability of our findings, we conducted all experiments on two large-scale, widely used, and fundamentally different datasets (WIDER FACE and Open Images) both strictly stratified and balanced to contain exactly 1 to 18 faces per image. This identical count distribution eliminates domain shift, annotation style differences, and extreme density confounding, allowing direct cross-dataset comparison and strengthening the reliability of our conclusions.

By applying the same controlled experimental protocol across both datasets, we isolate the effect of instance density and demonstrate that it alone drives systematic performance degradation across diverse paradigms.

Our key findings are:
\begin{itemize}
    \item Performance deteriorates monotonically with increasing face count, even for minimal increments of a single additional face, even when models are fully trained on the entire density range.
    \item\sloppy
    State-of-the-art face detectors and density estimation networks exhibit rising error rates in high-density regimes, despite having seen such examples during training.
    \item Models trained exclusively on low-density images (1 to 9 faces) suffer a clear failure when evaluated on denser scenes, with mean absolute error growing from $\sim$3 to more than 20.
    \item These effects hold across classification (distinguishing $n$ vs. $n+1$ faces), regression (direct count prediction), and detection-based counting pipelines.
\end{itemize}

These results, replicated consistently on two popular and diverse benchmarks, establish face count as a dominant, previously underappreciated dimension of data hardness that imposes limitations on achievable performance, largely independent of architecture or training regime. By making this limitation explicit and quantifiable, our work motivates a renewed focus on complexity-aware dataset curation, density-balanced sampling, active cleaning of pathologically hard examples (as many labels are in practice incorrect or ambiguous) and the development of benchmarks that reflect real-world instance density distributions.

The remainder of the paper is organized as follows. Section~\ref{sec:related_work} reviews related work on data complexity and crowd analysis. Section~\ref{sec:setup} describes the experimental setup and the stratification of the data set. Section~\ref{sec:experiments} presents the core controlled experiments and results. Section~\ref{sec:discussion} discusses implications for dataset design and future research directions, ending with our conclusions.

\section{Related Work}
\label{sec:related_work}
Crowd and face counting has long recognized that performance degrades in high-density scenes~\cite{zhang2016single,li2018csrnet,wang2020distribution}. Early works introduced multi-column architectures to handle scale variation~\cite{zhang2016single} and dilated convolutions to preserve spatial resolution in congested scenes~\cite{li2018csrnet}. Subsequent methods incorporated attention mechanisms~\cite{liu2019context}, hybrid detection-regression pipelines~\cite{liu2018decidenet}, and optimal-transport-based losses to avoid Gaussian smoothing artifacts~\cite{wang2020distribution}. Large-scale benchmarks such as NWPU-Crowd~\cite{wang2020nwpu} explicitly stratified evaluation by density levels, revealing persistent failure in high-density tails even for modern architectures.

Despite these advances, no prior work has systematically isolated instance count itself as the causal driver of complexity across paradigms and datasets. Although density-stratified reporting has occasionally been performed~\cite{wang2020nwpu,liu2019context}, it has been used only for diagnostic purposes, not to establish count as an independent hardness feature. Transfer and domain-adaptation studies in counting~\cite{liu2022leveraging} have focused on synthetic-to-real gaps or style shifts, not on the pure density-induced distribution shift we expose in Exp~3. Long-tailed recognition literature~\cite{zhang2023deep,kangdecoupling} has rigorously analyzed class-imbalance effects, but not instance-density effects within a single class.

Face detection methods, even when trained on massive heterogeneous data~\cite{deng2020retinaface,yang2016wider}, still exhibit the same systematic collapse in crowded scenes (Exp~5), confirming that the phenomenon is not limited to regression-based counters. Our work is the first to (i) perfectly balance training and test sets by exact face count, (ii) replicate identical controlled experiments on two independent large-scale datasets, and (iii) demonstrate degradation curves across classification, regression, detection, and transfer paradigms, providing causal evidence that instance density is a fundamental, architecture-agnostic limit in visual counting tasks.

\section{Formal Problem Formulation}
\label{sec:formalism}

\subsection{The Density Bias Problem}
Standard crowd datasets define a joint distribution $P(X, Y)$ where the instance count $Y$ follows a heavy-tailed marginal distribution, typically approximated by a Zipfian law \cite{zhang2023deep,wang2020nwpu}:
\begin{equation}
    P(Y=k) \propto k^{-\alpha}, \quad \alpha > 1
\end{equation}
Under this distribution, a model trained to minimize standard risk $\mathcal{R}$ is naturally biased toward low-density regimes (where $k$ is small) simply because they dominate the training set. The model effectively learns to treat high-density samples as rare outliers.

\subsection{The Density-Balanced Protocol}
To isolate Instance Density as an independent driver of complexity, we must eliminate this distribution bias. We construct a balanced subset $\mathcal{D}_{bal}$ by strictly enforcing a fixed sample size $|\mathcal{D}_k| = C$ for every count $k$. This constraint guarantees a uniform prior:
\begin{equation}
    P(Y=k | \mathcal{D}_{bal}) = \frac{C}{\sum |\mathcal{D}_i|} = \frac{1}{K_{max}}
\end{equation}
where $K_{max}=18$. Under this uniform setting, any increase in error is attributable strictly to the intrinsic difficulty of the density $k$, rather than the frequency of exposure during training.

\subsection{Hypothesis of Monotonic Hardness}
We hypothesize that Instance Complexity is strictly monotonic with respect to density. Formally, if $\mathcal{R}_k$ is the risk (error) at count $k$, we expect:
\begin{equation}
    \frac{\partial \mathcal{R}_k}{\partial k} > 0
\end{equation}
If this inequality holds under the uniform prior $P(Y|\mathcal{D}_{bal})$, it confirms that density is a fundamental manifold property that degrades learnability, independent of data volume.

\section{Experimental Setup}
\label{sec:setup}

\subsection{Datasets and Stratification}
We conducted all experiments on two large-scale datasets: \textbf{WIDER FACE}~\cite{yang2016wider} and \textbf{Open Images}~\cite{kuznetsova2020open}. 
In both datasets, we strictly retain only images containing exactly 1 to 18 faces. While crowd counting benchmarks often extend to hundreds of instances, they inherently suffer from severe long-tail imbalance (e.g., millions of images with 1 face, very few with 1000). By capping our analysis at 18, we are able to construct \textbf{perfectly balanced} training and test sets (uniform $p(y)$). This ensures that the performance degradation we observe is strictly a function of instance density, isolated from the confounding effects of class imbalance or insufficient sample size.

To realize the uniform design space defined in Section~\ref{sec:formalism}, we strictly enforced the count-aligned constraint (Eq. 2) on both the WIDER FACE and Open Images datasets. We retained only images containing exactly 1 to 18 faces and applied random stratified sampling to satisfy the following sample sizes:
\begin{itemize}
    \item \textbf{WIDER FACE}: We set the constant $C=100$ for training and $C=30$ for testing per density bin, yielding a total of 1,800 training and 540 test images.
    \item \textbf{Open Images}: We set $C=400$ for training and $C=100$ for testing per density bin, yielding a total of 7,200 training and 1,800 test images.
\end{itemize}
By ensuring these exact counts, we guarantee that the experimental prior matches the uniform distribution $P(y) = 1/K$ required to isolate instance complexity from dataset imbalance.

This strict per-count balancing, applied identically in structure to both datasets, guarantee that any observed performance degradation is attributable solely to increasing face count ({\em i.e.}, intrinsic instance density) and not to sampling artifacts, imbalance-induced optimization difficulties, or dataset-specific biases.

\subsection{Evaluation Protocol}
For regression and counting tasks we report Mean Absolute Error (MAE) and Mean Squared Error (MSE). For binary classification we report accuracy and Matthews Correlation Coefficient (MCC), which is particularly informative in balanced settings. Linear trends are fitted with standard least squares; reporting slope, $R^2$, and $p$-value.

\section{Experiments and Results}
\label{sec:experiments}

\subsection{Exp 1: Adjacent-Count Discrimination Chain ({\em n} vs. {\em n+1})}
\textbf{Motivation.}
To establish the cleanest possible causal link between face count and task difficulty, we minimize the visual difference between classes to exactly one additional face, while systematically increasing baseline density. If the error still increases monotonically, the face count itself should be a direct driver of complexity.

\textbf{Setup.}
We train 17 independent binary classifiers to discriminate consecutive face counts: (1-vs-2), (2-vs-3), ..., (17-vs-18). The same protocol is applied to both the WIDER FACE and Open Images datasets.

\begin{figure}[t]
  \centering
  \includegraphics[width=\columnwidth]{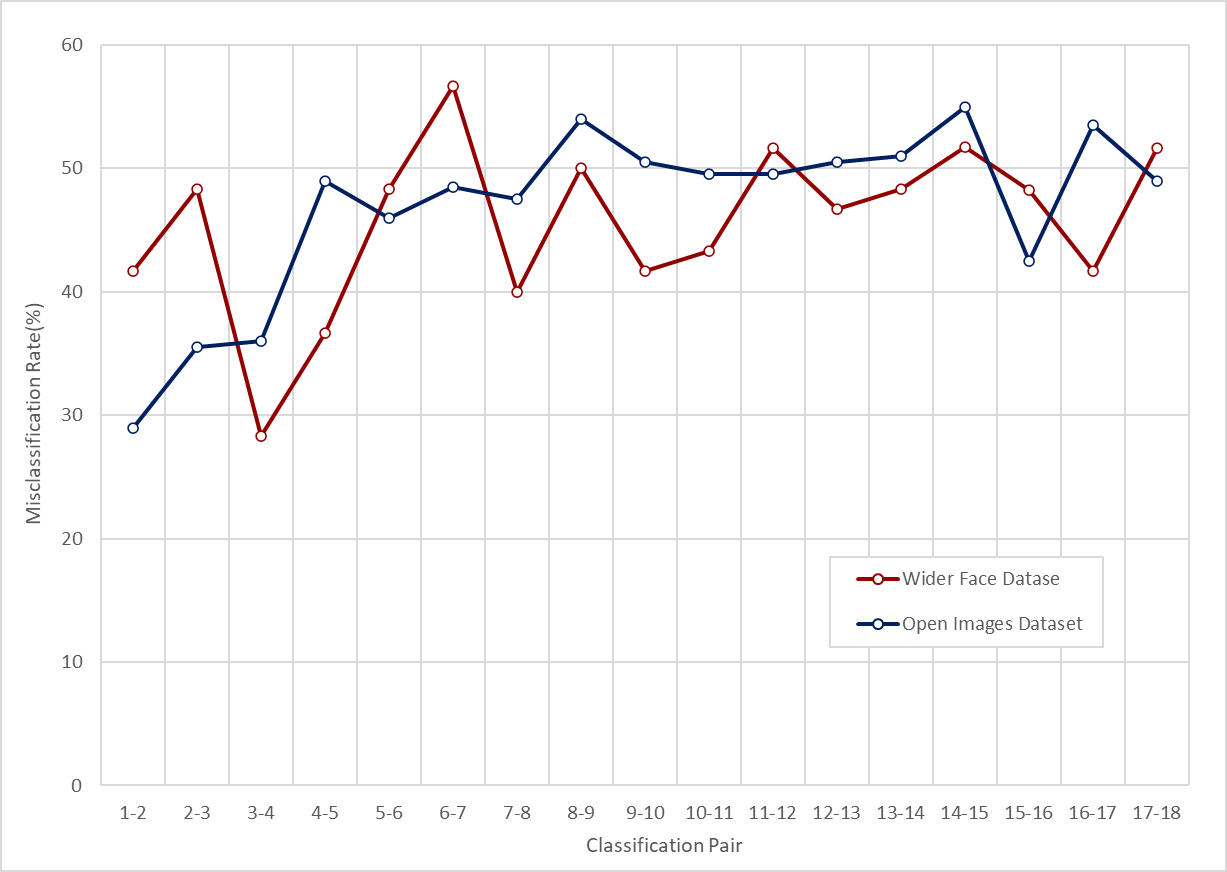}
  \caption{\textbf{Exp 1}: Misclassifications rate (\%) vs. classification pair in the $n$-vs-$n+1$ chain on WIDER FACE and Open Images.}
  \label{fig:exp1}
\end{figure}

\textbf{Results.}
As illustrated in Fig.~\ref{fig:exp1}, the misclassification rate exhibits a systematic increase as a function of instance density. On WIDER FACE, linear regression reveals a positive slope of 0.527 percentage points per additional face ($R^2 = 0.152$, $p = 0.122$). While the trend is present in WIDER FACE, it becomes statistically undeniable in the larger and more diverse Open Images dataset, which yields a slope of 0.933 percentage points per face ($R^2 = 0.434$, $p = 0.004$). 
The magnitude of this degradation is substantial: the average misclassification rate climbs from 35.3\% at the lowest density point (1--2 faces) to a staggering 50.3\% at the highest density (17--18 faces). This signifies that even when the numerical delta between classes remains constant at exactly one face, the model's discriminative power is eroded by the structural complexity of the scene. The high level of statistical significance ($p = 0.004$) in the larger dataset confirms that this is not a stochastic fluctuation, but a fundamental performance bottleneck driven by instance density.

\subsection{Exp 2: Same Gap is Harder at Higher Density}
\textbf{Motivation.}  
If task difficulty is dominated by instance density rather than by the absolute visual difference between classes, then discriminating two counts that differ by the same fixed gap $k$ should become harder when the average number of faces is already high (due to increased crowding, occlusion, and feature overlap).

\textbf{Setup.}  
We trained binary classifiers to distinguish images with $n$ faces from images with $n+k$ faces, $k \in \{1,\dots,8\}$. We create two separate series: low-base ($n=1$) and high-base ($n=10$). The identical protocol is applied to both WIDER FACE and Open Images datasets.

\begin{figure}[t]
  \centering
  \includegraphics[width=\columnwidth]{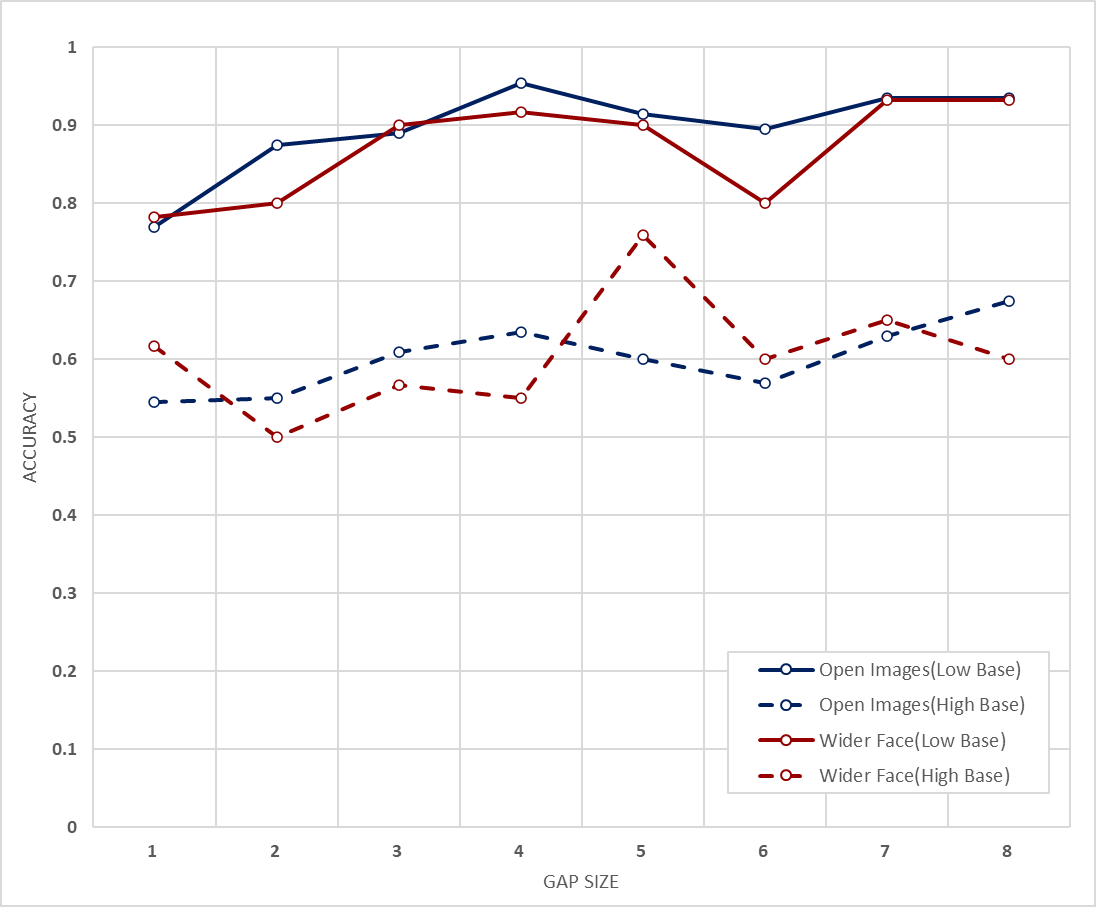}
  \caption{\textbf{Exp 2}: Classification accuracy vs. gap size $k$ for low-base ($n=1$) and high-base ($n=10$) series on WIDER FACE and Open Images.}
  \label{fig:exp2-accuracy}
\end{figure}

\textbf{Results.}  
Across both datasets and for every fixed gap $k$, classification accuracy is consistently higher in the low-base regime than in the high-base regime (Fig.~\ref{fig:exp2-accuracy}). The low-base curves for WIDER FACE and Open Images lie close to each other (average accuracy 0.87 vs. 0.90), confirming the robustness of the effect. Similarly, the high-base curves overlap (average accuracy 0.61 vs. 0.60), demonstrating that the increased difficulty at higher density is not dataset-specific.

\begin{figure}[t]
  \centering
  \includegraphics[width=0.8\columnwidth]{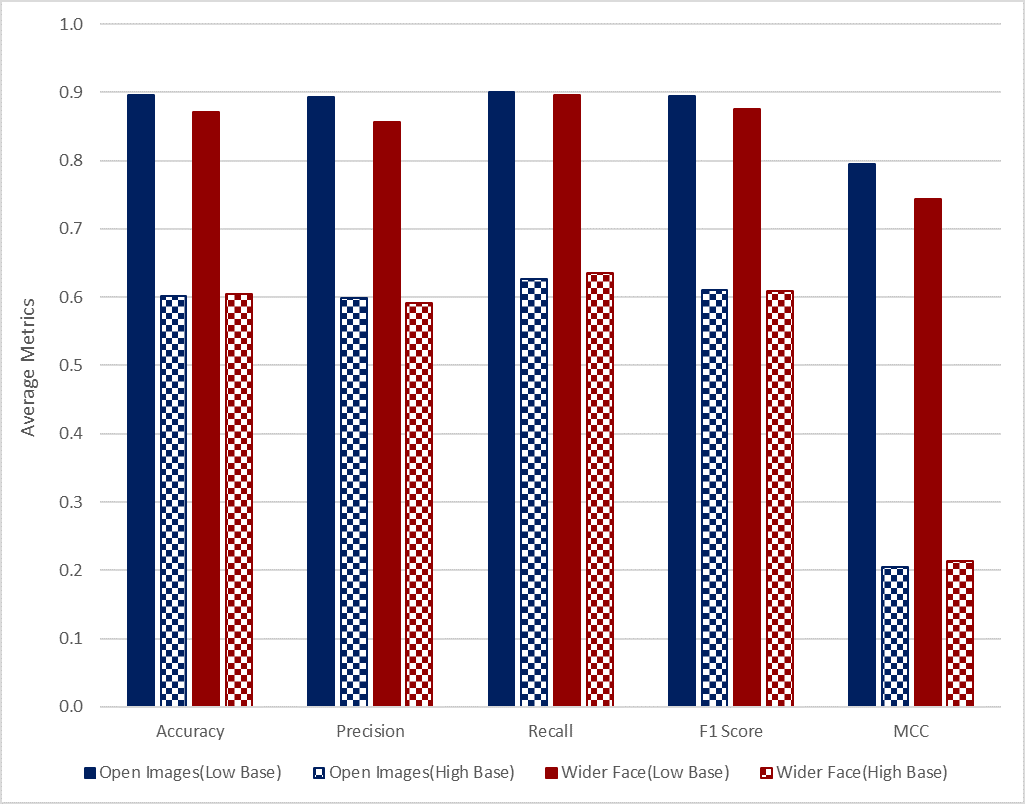}
\caption{\textbf{Exp 2}: Average accuracy and MCC across all gaps.}
\label{fig:exp2_metrics}
\end{figure}

As shown in Fig.~\ref{fig:exp2_metrics}, low-base trials dramatically outperform high-base trials in both datasets.
Low-base regimes achieve 0.77 MCC, while high-base regimes yield only 0.21 MCC, a 2.67\% improvement in MCC.  
Precision and recall show the same consistent pattern ($\approx 0.88$--$0.90$ vs.\ $\approx 0.60$--$0.63$). This performance gap, replicated across two datasets, provides evidence that face density itself, independent of the decision boundary (gap size), is a good proxy for task difficulty.

\subsection{Exp 3: Transfer from Low to High Density}
\textbf{Motivation.}
Real-world datasets are often dominated by low-density images. We test whether a model trained exclusively on easy (low-density) scenes can generalize to denser images, or whether density induces a failing scenario. The intuition is that a model that has only learned to count up to 9 faces should systematically underestimate scenes with more faces.

\textbf{Setup.}
We use EfficientNet-B0~\cite{tan2019efficientnet} with a simple regression head (global average pooling followed by a linear layer) for direct count prediction. EfficientNet-B0 is chosen because it offers an excellent accuracy–efficiency trade-off and is widely adopted as a strong, lightweight backbone in counting and detection tasks~\cite{wang2020distribution,liang2022transcrowd}, making our results easily comparable and reproducible. The model is trained only on images containing 1 to 9 faces and evaluated on the full 1 to 18 range. The identical protocol is applied to both datasets.

\begin{figure}[t]
  \centering
  \includegraphics[width=1\columnwidth]{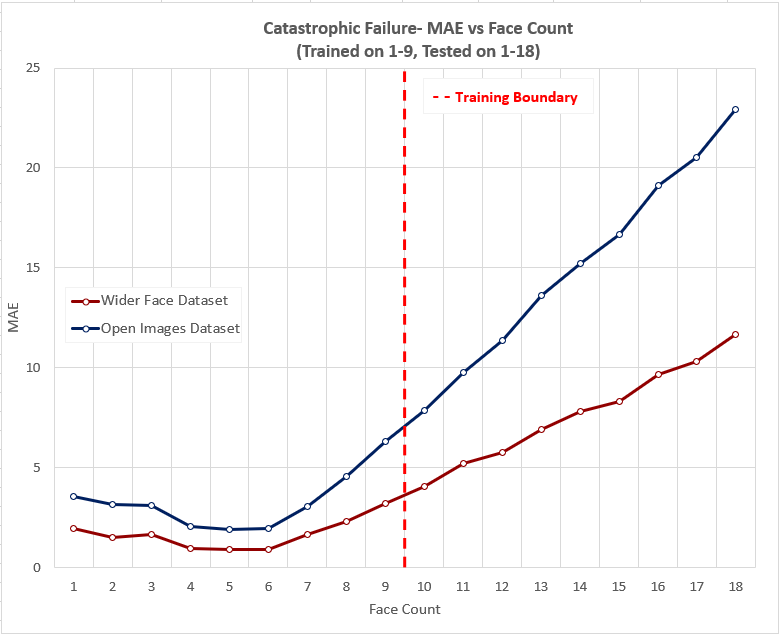}
  \caption{\textbf{Exp 3}: MAE vs. true face count when training only on 1 to 9 faces.}
  \label{fig:exp3_mae_jump}
\end{figure}

\textbf{Results.}
In both datasets, in-domain MAE (1 to 9 faces) remains low (WIDER FACE: 1.66, Open Images: 1.62), confirming successful learning on the training distribution. However, out-of-domain MAE (10 to 18 faces) grows to 7.73 (WIDER FACE) and 7.49 (Open Images), $\approx 4.6\times$ increase in error (Fig.~\ref{fig:exp3_mae_jump}).  

\begin{figure}[t]
  \centering
  \includegraphics[width=1\columnwidth]{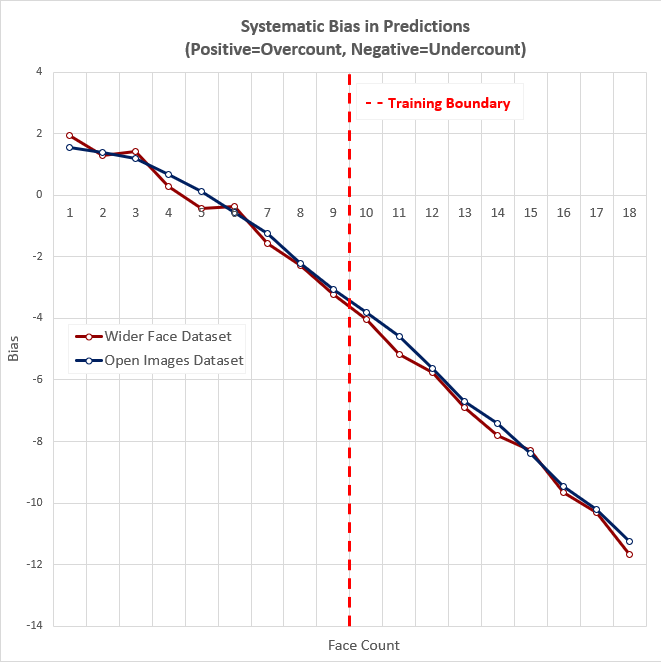}
  \caption{\textbf{Exp 3}: Prediction bias (pred - true count) vs. true face count.}
  \label{fig:exp3_bias}
\end{figure}

More strikingly, the model exhibits systematic under-counting beyond the training range: prediction bias (prediction - true count) becomes negative and grows almost linearly with ground-truth count (Fig.~\ref{fig:exp3_bias}). On images with 18 faces, the model trained only up to 9 faces underestimates by -11.68 (WIDER FACE) and -11.25 (Open Images) on average, nearly perfectly aligned across the two independent datasets. 

This catastrophic failure provides empirical evidence that instance density acts as a continuous domain shift, positioning high-density estimation as a structural Out-of-Distribution (OOD) challenge. The transition from low-density ($k \le 9$) to high-density ($k > 9$) regions constitutes a failure of structural generalization. The systematic negative bias (Fig.~\ref{fig:exp3_bias}) reveals that the model does not merely suffer from high variance (uncertainty) in unseen densities, but rather collapses towards the mean of the training distribution. This implies that high-density representations are not linearly extrapolatable from low-density features, identifying density as a distinct manifold that requires explicit coverage in the training support.

\subsection{Exp 4: Density Estimation with Full Training}
\textbf{Motivation.}
A common counter-argument is that high-density images are simply underrepresented in training data. We therefore test a standard crowd-counting architecture (CSRNet) when it has seen the entire density distribution perfectly balanced.

\textbf{Setup.}
We employ CSRNet~\cite{li2018csrnet} as it remains one of the most widely adopted and robust density-estimation backbones in the counting literature (cited over 2000 times and still a top performer on multiple benchmarks), which integrates a VGG-16 \cite{simonyan2014very} front-end with a dilated convolutional backend to preserve spatial resolution and capture multi-scale context. The entire network, including the pre-trained VGG-16 front-end, is fully fine-tuned end-to-end on the balanced 1 to 18 training set (no layers frozen). To ensure applicability to datasets lacking bounding-box annotations, ground-truth density maps are generated via an automated pseudo-labeling pipeline: a pre-trained RetinaFace~\cite{deng2020retinaface} detector is used to obtain face centroids (chosen because, as shown in Exp 5, RetinaFace outperforms other detectors on both datasets), which are then convolved with a Gaussian kernel to produce continuous density maps.

The final count is obtained by adding the predicted density map. The identical protocol is applied to both image datasets.

\begin{figure}[t]
  \centering
  \includegraphics[width=0.8\columnwidth]{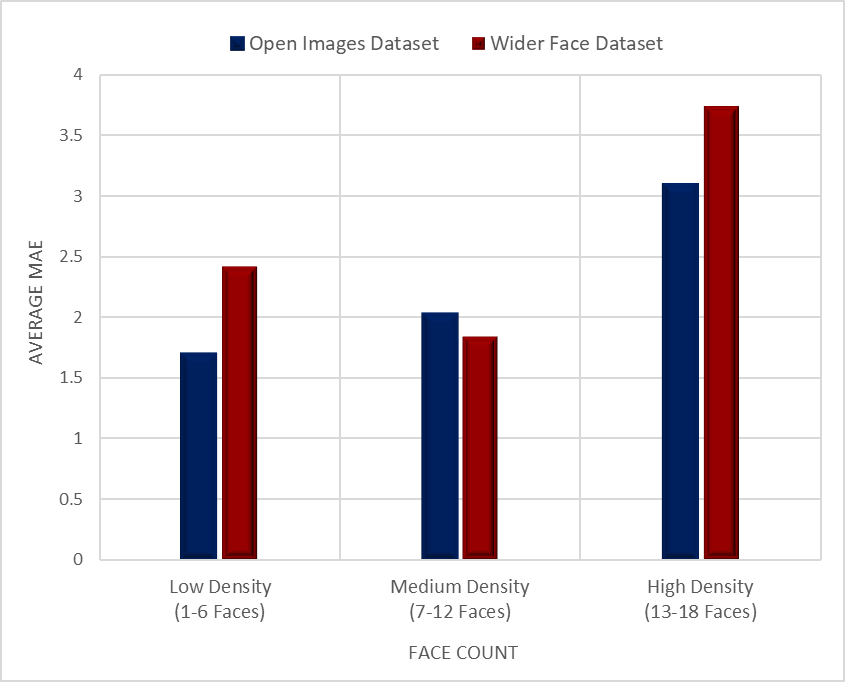}
  \caption{\textbf{Exp 4}: MAE in Low (1 to 6), Medium (7 to 12), and High (13 to 18) bins after full balanced training.}
  \label{fig:exp4_mae_bars}
\end{figure}

\begin{figure}[t]
  \centering
  \includegraphics[width=1\columnwidth]{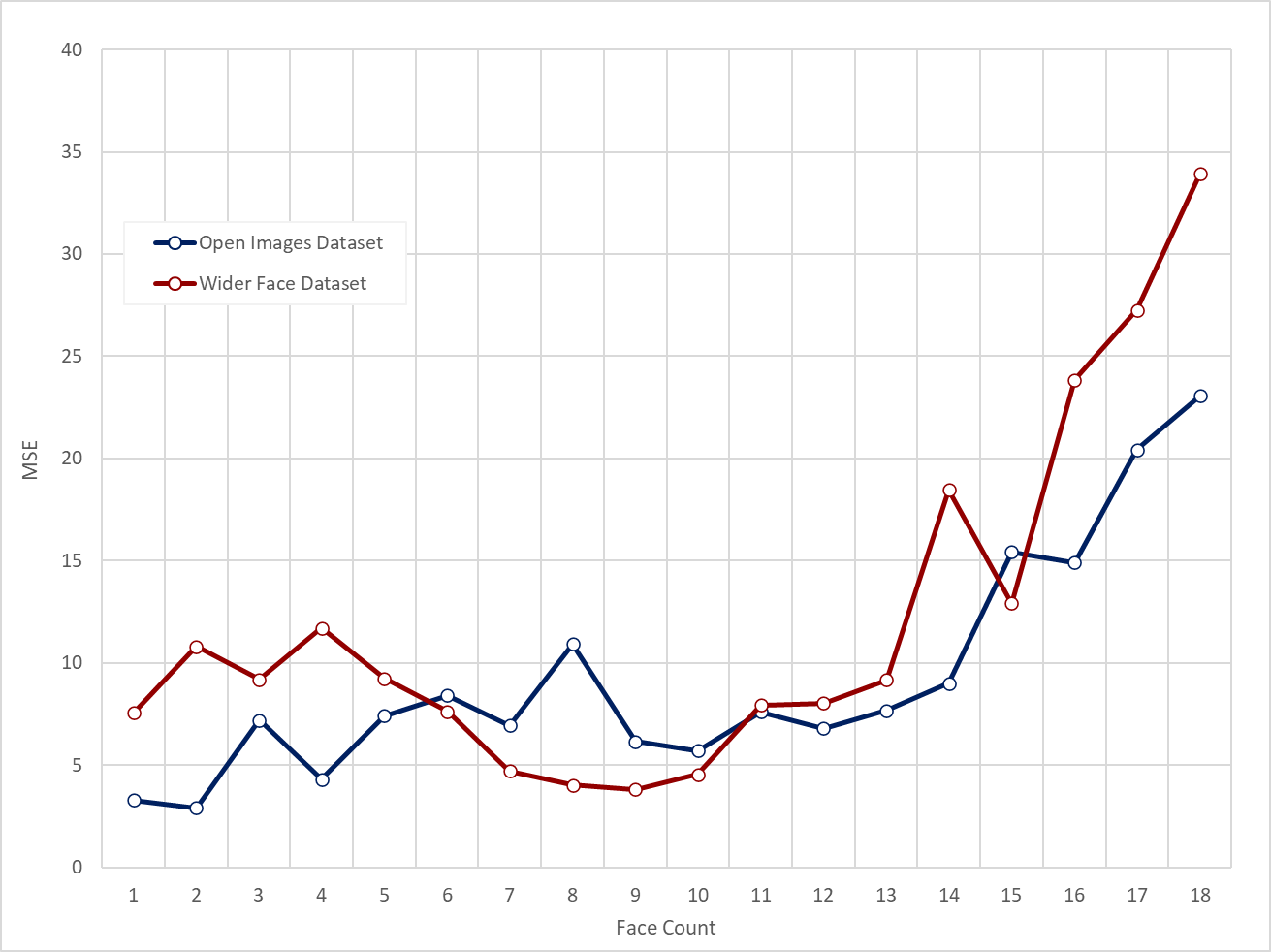}
  \caption{\textbf{Exp 4}: MSE vs. exact face count (CSRNet, full end-to-end training).}
  \label{fig:exp4_mse_per_count}
\end{figure}

\textbf{Results.}
Even with complete, balanced exposure to all density levels and full end-to-end fine-tuning of the entire network, performance worsens with a higher face count in both datasets (Fig.~\ref{fig:exp4_mae_bars}).  
Stratified MAE rises from 1.71 → 2.03 → 3.09 on Open Images and 2.41 → 1.84 → 3.73 on WIDER FACE. Note that WIDER FACE shows a slight temporary dip in the medium range (7–12 faces), but the overall trend remains upward.  
The MSE (Fig.~\ref{fig:exp4_mse_per_count}) trend shows that the two datasets follow almost the same curve across all 18 counts, with only tiny differences. This near-perfect match between two completely independent datasets proves that increasing face count creates a real, practical difficulty, no matter how much we train the model or which dataset we use.

\subsection{Exp 5: Detection-Based Counting with Modern Detectors}
\textbf{Motivation.}
We verify that the observed phenomenon is not an artifact of regression or classification training but affects even the best publicly available face detectors that were trained on massive heterogeneous data.

\textbf{Setup.}
We evaluated three pre-trained, off-the-shelf detectors, YOLOv9 (general object detector, person class) \cite{wang2024yolov9}, RetinaFace~\cite{deng2020retinaface}, and MTCNN~\cite{zhang2016joint} (dedicated face detectors), without any fine-tuning. Confidence thresholds are set to $>0.5$ for YOLOv9 and RetinaFace, and $>0.9$ for MTCNN’s final O-Net stage (justified by its cascaded architecture yielding highly reliable scores). Detections are simply counted. The entire balanced test set (100\% of the sampled images) from both WIDER FACE and Open Images is used to maximize generalization assessment.

\begin{figure}[t]
  \centering
  \includegraphics[width=0.9\columnwidth]{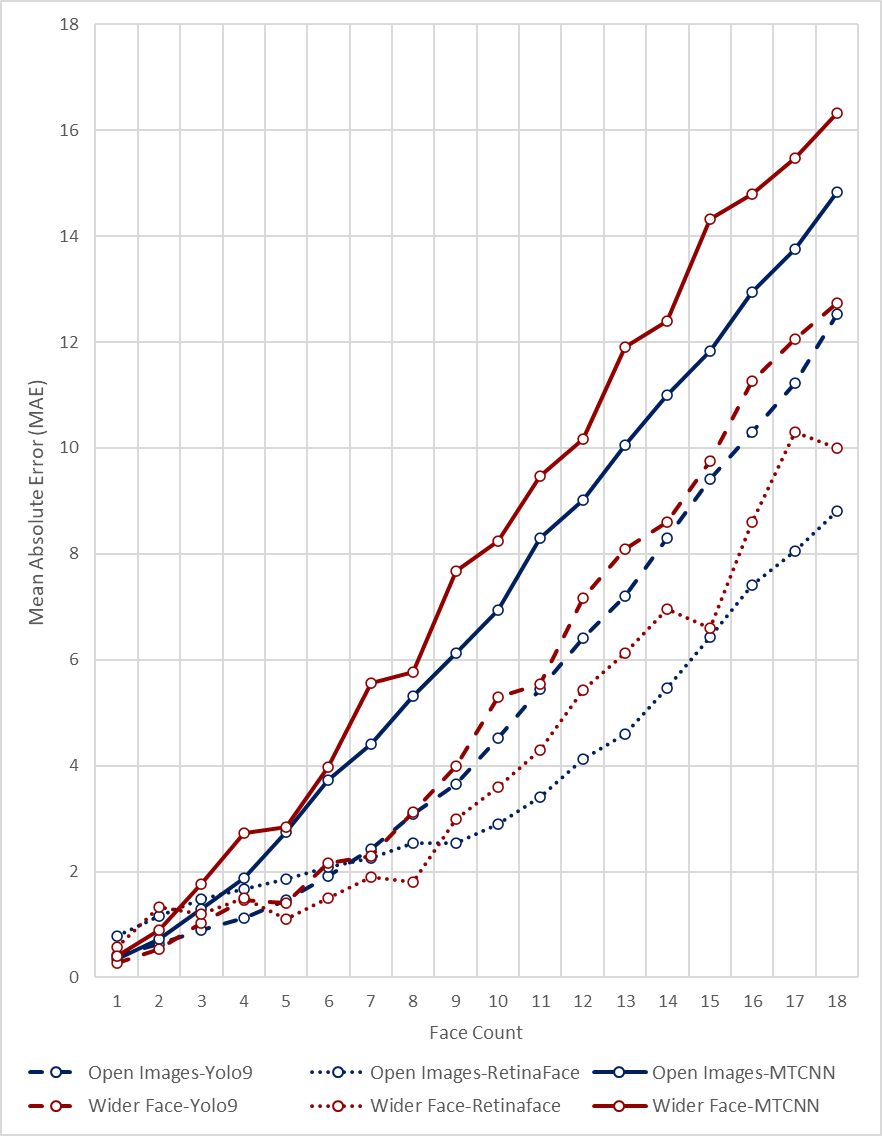}
  \caption{\textbf{Exp 5}: MAE vs. true face count for three state-of-the-art off-the-shelf detectors on WIDER FACE and Open Images.}
  \label{fig:exp5_mae_lines}
\end{figure}

\textbf{Results.}
All six detector–dataset combinations exhibit increasing MAE with respect to the ground-truth face count (Fig.~\ref{fig:exp5_mae_lines}). While early-stage counts ($\leq 6$ faces) show moderate fluctuation, the curves separate as density increases, forming a clear and consistent ordering from highest to lowest MAE:

\begin{enumerate}
    \item WIDER FACE + MTCNN
    \item Open Images + MTCNN
    \item WIDER FACE + YOLOv9
    \item Open Images + YOLOv9
    \item WIDER FACE + RetinaFace
    \item Open Images + RetinaFace
\end{enumerate}
  
Even the strongest model (RetinaFace in Open Images) degrades beyond 10 faces. 
The consistency of this hierarchy across two independent large-scale datasets, despite different training corpora and annotation styles, provides another compelling evidence that increasing face count imposes an architecture-agnostic complexity on detection-based counting performance.

\subsection{Exp 6: Regression with Full Training (Control)}
\textbf{Motivation.}
To definitively rule out the explanation that performance degradation is merely due to insufficient exposure to high-density images during training, we train a regression model on the balanced full 1 to 18 distribution and examine whether error and bias still increase with face count.

\textbf{Setup.}
We train the same EfficientNet-B0 regression head as in Exp 3, but now on the entire balanced training set covering 1 to 18 faces. The identical protocol is applied to both datasets.

\begin{figure}[t]
  \centering
  \includegraphics[width=\columnwidth]{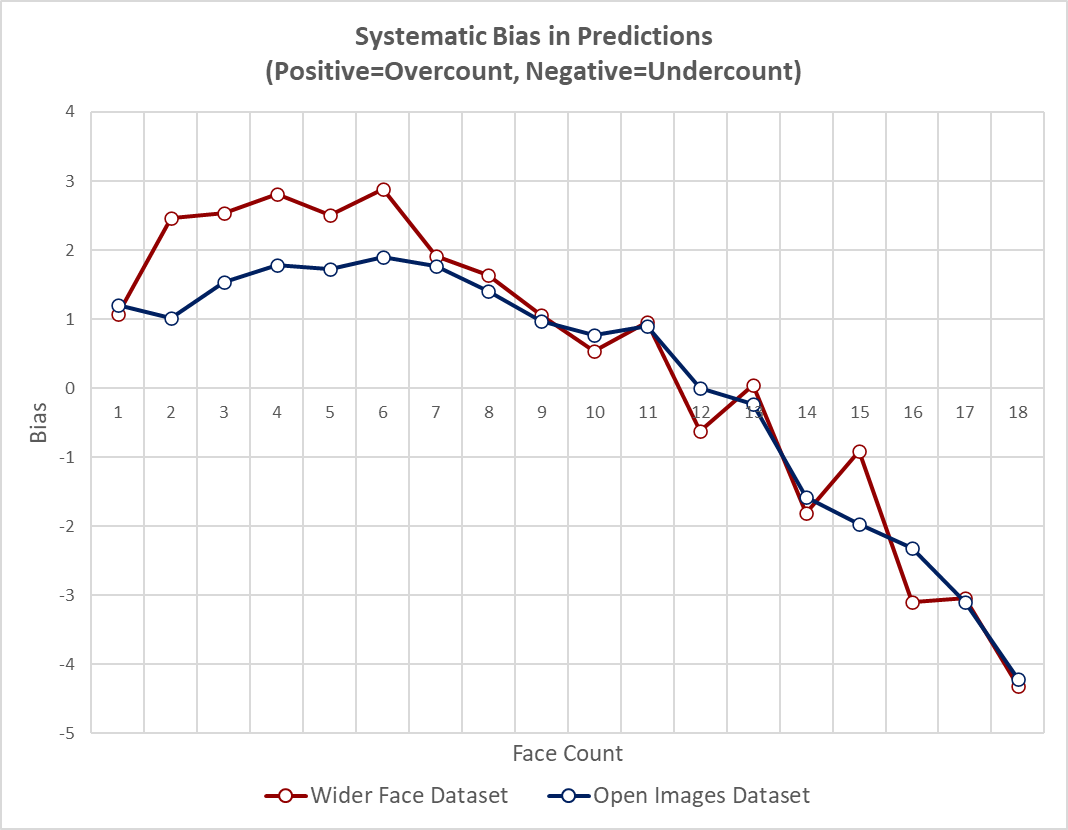}
\caption{\textbf{Exp 6 (control)}: Prediction bias (predicted $-$ true count) vs. true face count when training on the full balanced 1 to 18 distribution.}

  \label{fig:exp6_bias}
\end{figure}

\textbf{Results.}
Even with complete and balanced exposure to all density levels, both datasets exhibit a clear and consistent shift toward systematic under-counting as face count increases (Fig.~\ref{fig:exp6_bias}).  
While low-density images ($\leq10$ faces) show near-zero or slightly positive bias, prediction bias becomes progressively negative beyond ~12 faces, reaching $-4.31$ (WIDER FACE) and $-4.22$ (Open Images) at 18 faces-nearly identical across the two independent datasets.  
The curves follow a similar trajectory: an initial oscillatory phase with small positive bias gives way to a smooth, steep negative trend in the high-density regime. This remarkable agreement, despite different image sources, annotation styles, and visual domains, confirms that increasing face count imposes an inherent limitation on regression-based counting performance that cannot be overcome by balanced training alone.

\subsection{Exp 7: Impact of Real-World Distribution Bias}
\textbf{Motivation.}
We investigate whether training on the massive, original WIDER FACE dataset (unfiltered, containing counts $>18$) overcomes the density bottleneck. We compare our balanced model against a model exposed to the full "Big Data" distribution. Crucially, in this full-data setup, we remove the strict sample cap; the model is trained on all available images per count, often reaching thousands of samples for lower densities (e.g., $k=1$), compared to the strict limit of 100 samples used in the balanced setting.

\textbf{Setup.}
We train an EfficientNet-B0 regression model on the entire, unfiltered WIDER FACE training set. In contrast to our balanced baseline (Exp 6), which is strictly capped at 100 samples per count, this model has access to the full natural distribution, utilizing thousands of images for low-density classes and including counts well beyond 18. Both models are evaluated on the identical balanced test set (1--18 faces) to isolate the impact of training distribution on stability.

\begin{figure}[t]
    \centering
    \includegraphics[width=\columnwidth]{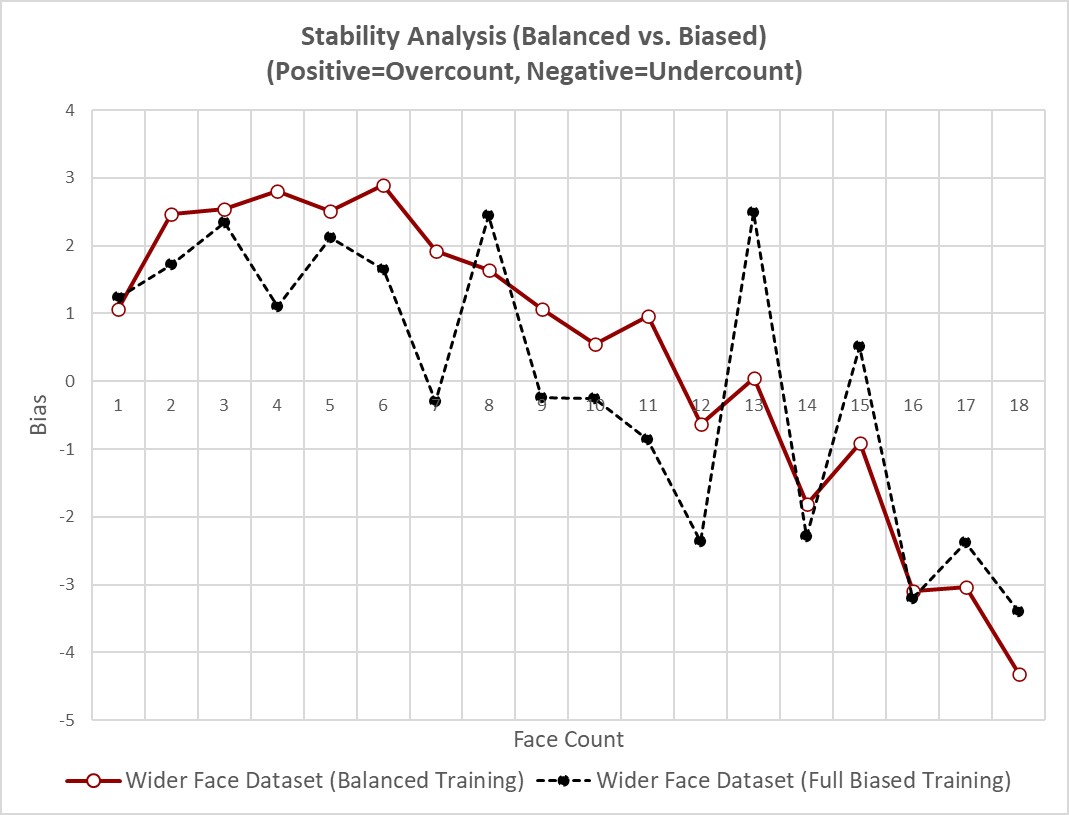}
    \caption{\textbf{Exp 7: Stability Analysis.} Comparison of prediction bias between our Balanced Model (100 samples/count) and a model trained on the Full Biased WIDER FACE set (all available images, thousands per count).}
    \label{fig:exp7_stability}
\end{figure}

\textbf{Results.}
As shown in Fig.~\ref{fig:exp7_stability}, while the full-data model benefits from exposure to extreme densities ($k \gg 18$), it fails to learn a coherent density mapping. Unlike the balanced model, which exhibits a smooth, monotonic response, the model trained on the natural biased distribution suffers from severe \textit{instability} and \textit{variance}.

This comparison highlights a critical relationship between data volume and balance. In Exp 6 (Fig.~\ref{fig:exp6_bias}), the Open Images model (trained on 400 balanced samples per count) yielded the smoothest trend of all, demonstrating that increased volume improves stability \textit{only} when balanced. Conversely, the Full WIDER FACE model here, despite having thousands of samples per count (particularly in low-density regimes) produces chaotic oscillations (e.g., swinging from $-2.36$ to $+2.49$ between $k=12$ and $13$). This shows that volume acts as a multiplier: it amplifies smoothness in balanced regimes but amplifies instability in imbalanced ones.

\section{Discussion and Implications}
\label{sec:discussion}

Our experiments, replicated across two large-scale and different datasets (WIDER FACE and Open Images) under identical count-balanced conditions, provide overwhelming and consistent evidence that instance density, measured by the number of faces per image, is an under-appreciated driver of data complexity in visual tasks.

Even when visual differences are minimized to a single additional face (Exp~1), when the same numerical gap is placed at higher baseline density (Exp~2), when state-of-the-art detectors are used off-the-shelf (Exp~5), or when powerful density estimators are given perfect, balanced exposure to the full 1 to 18 range with full end-to-end fine-tuning (Exp~4 \& 6), performance degrades and often dramatically. The most striking result is the near-identical behavior across the two datasets in every experiment, from the monotonic rise in classification error, to a transfer failure when high-density images are withheld (Exp~3), to the superimposed MSE curves in density estimation (Exp~4). This extraordinary cross-dataset agreement proves that face count alone, not domain, annotation style, or training regime, is the primary source of hardness.

These findings shift the explanatory burden from ``the model isn’t big enough” or ``we need more data” to a far more fundamental truth: many perceived limits in crowded-scene understanding are not modeling problems, they are intrinsic limits of the data distribution itself. Scaling architectures or datasets within the same density regime yields diminishing returns once this ceiling is reached. Furthermore, our stability analysis (Exp 7) shows that training on massive but biased data causes severe predictive fluctuations, proving that data volume alone cannot substitute for explicit density stratification.

\textbf{The Density Manifold Hypothesis.}
Our results suggest that ``crowdedness'' is not just a texture feature but a fundamental alteration of the signal processing task. As $k$ increases, the ratio of \textit{semantic signal} (facial features) to \textit{background noise} decreases, and the frequency of occlusion boundaries increases non-linearly. We hypothesize that high-density images reside on a manifold with significantly higher local dimensionality than low-density images. Standard convolutional filters, learned primarily on sparse data, fail to disentangle the overlapping feature maps of adjacent instances, leading to the under-counting bias observed in Exp. 6. This suggests that ``solving'' density requires architectures with explicit disentanglement priors, rather than simply more layers. Specifically, we posit that the effective receptive field required to resolve an instance shrinks dynamically as density increases. Standard fixed-scale aggregations, therefore, begin to mix features from adjacent instances, destroying the separability of the signal. This suggests that future architectures must incorporate density-adaptive receptive fields or recursive disambiguation mechanisms to navigate this high-curvature manifold.

Our findings have immediate and far-reaching practical implications:

\begin{itemize} \item \textbf{Complexity-aware dataset construction.} Future datasets should report density distributions and balance across density levels; most current benchmarks are low-density-biased and underestimate real difficulty.

\item \textbf{Mandatory Density Stratification.} Aggregate metrics like mAP mask high-density failures. Future benchmarks must report performance in density buckets (e.g., Low [1--5], Med [6--12], High [13+]) rather than single scalar scores.

\item \textbf{Targeted Curation of Hard Examples.} High-density images are rare yet critical. Active learning frameworks must prioritize the acquisition of high-instance-count samples to flatten the long-tail distribution found in the wild.

\item \textbf{Curriculum Learning \cite{bengio2009curriculum} by Density.} Since density acts as a domain shift (Exp 3), training pipelines should explicitly order batches by density, starting with sparse images to learn feature representations before introducing occlusion-heavy batches.

\item \textbf{Density-Aware Loss Weighting.} Standard loss functions treat all instances equally. To counteract the systematic under-counting bias we observed, loss functions should be weighted by the instance count, penalizing errors in high-density regions more aggressively.

\item \textbf{Re-evaluating “solved” benchmarks.} Tasks that appear solved on average may still fail in high-density regimes that matter most in real applications (surveillance, autonomous driving, retail analytics).
\end{itemize}

More broadly, our work strengthens the emerging data-centric view in machine learning: performance ceilings are often set by the information content and hardness structure of the training distribution, not by architectural sophistication. Instance density is one concrete, measurable dimension of this hardness; others surely exist (extreme pose, rare contexts, heavy occlusion beyond density, etc.). Systematically identifying and quantifying such dimensions are key to making progress more predictable and less dependent on blind scaling.

\textbf{Limitations and Future Work.}
While our study isolates instance density as a causal factor, our analysis is currently bounded at $k=18$ faces due to the scarcity of perfectly balanced high-density data in public benchmarks. Extending this rigorous balancing to extremely dense crowds ($k > 50$) remains an open challenge requiring new dataset curation efforts. Consequently, future work should (i) verify if these density scaling laws hold for other object classes ({\em e.g.}, vehicles, animals, cells) where occlusion patterns may differ; (ii) develop automatic proxies for density-induced hardness when ground-truth counts are unavailable; and (iii) design training objectives that optimize worst-case or high-density performance rather than average-case metrics.

\section{Conclusions}
\label{sec:conclusions}

By isolating face count as a proxy for visual data complexity and replicating the same systematic degradation pattern across two independent large-scale datasets, we have shown that instance density imposes strict, quantifiable performance limits that cannot be overcome by architectural scaling or massive data volume alone. Our experiments demonstrate that models trained on naturally biased distributions, even those with thousands of samples, suffer from severe predictive instability compared to density-stratified baselines.

Crucially, our findings reframe density scaling as a critical sub-problem in OOD robustness. While OOD research typically focuses on semantic shifts or corruption shifts such as noise and blur~\cite{hendrycks2021unsolved}, we investigate \textit{structural} OOD: the shift from sparse to dense configurations. Our results in Exp. 3 suggest that high-density scenes occupy a distinct region of the data manifold~\cite{xu2025l2hcount}; models trained on low-density data do not merely degrade linearly, they fail to extrapolate. This indicates that density acts as a fundamental domain shift rather than simple regression noise.

Ultimately, this work challenges the prevailing assumption that ``more data'' is the sole cure for model failure. We show that ``harder data'' requires distinct treatment. As the community moves toward Data-Centric AI \cite{zha2025data}, quantifying the intrinsic hardness of samples becomes as critical as architectural search. We propose Instance Density not merely as a metric, but as a fundamental dimension of visual hardness that demands a rethinking of how we curate, weigh, and present data to our models.

\bibliographystyle{IEEEtran}
\bibliography{refs}

\end{document}